\documentclass[10pt,twocolumn,letterpaper]{article}

\usepackage{cvpr}              
\usepackage{multirow}

\definecolor{cvprblue}{rgb}{0.21,0.49,0.74}
\usepackage[pagebackref,breaklinks,colorlinks,allcolors=cvprblue]{hyperref}

\def\confName{CVPR}
\def\confYear{2026}

\title{Translating Signals to Languages for sEMG-Based Activity Recognition}

\author{
Ming Wang$^{1}$ \quad
Haoxuan Qu$^{1}$\thanks{Corresponding Author}\quad
Qiuhong Ke$^{2}$ \quad
Wei Zhou$^{3}$ \quad
Hossein Rahmani$^{1}$ \quad
Jun Liu$^{1}$\\
$^{1}$Lancaster University \quad
$^{2}$Monash University \quad
$^{3}$Cardiff University\\
{\tt\small \{m.wang36,h.qu5,h.rahmani,j.liu81\}@lancaster.ac.uk,}\\
{\tt\small qiuhong.ke@monash.edu,\quad zhouw26@cardiff.ac.uk}\\
}

\begin{document}
\maketitle
\begin{abstract}
Surface electromyography (sEMG) signal-based activity recognition has attracted increasing research attention in recent years. To develop accurate sEMG signal-based activity recognizers, numerous approaches have been proposed. Some studies focus on designing larger and more expressive model architectures to enhance the representational capacity of sEMG signals, while others aim to enrich model priors through large-scale pretraining, thereby improving recognition performance. Recently, large language models (LLMs) have shown remarkable generalization and reasoning capabilities in natural language processing, whose implicit knowledge, learned from extensive linguistic descriptions of actions, opens new possibilities for interpreting sEMG signals and inferring activity intentions. Motivated by this, we propose LLM-sEMG, a novel framework that leverages LLMs as sEMG activity recognizers. Within this framework, we design a language-oriented mapping mechanism that converts continuous sEMG sequences into ``sEMG language,'' integrating several strategies to further facilitate the signal-to-language mapping process. Extensive experiments demonstrate that the proposed framework achieves highly accurate sEMG signal-based activity recognition using large language models.
\end{abstract}
    
\section{Introduction}
\label{sec:intro}

The core of activity recognition lies in accurately inferring and classifying human motion intentions from physiological or sensor-derived signals. This task is of fundamental importance in domains such as human-computer interaction, multimodal human activity understanding, and embodied perception~\cite{Ci2023CVPR,Zhang2024CVPR,Buchner2025CVPR,Chen2021SensorHAR}. As a direct reflection of neuromuscular activity, surface electromyography (sEMG) provides high temporal resolution and captures subtle dynamic variations in muscle activation, offering rich temporal features for movement intention decoding~\cite{Farina2025sEMG,Ma2025}. As a result, sEMG-based activity recognition has attracted substantial research interest in recent years~\cite{Zhang2025EMGArmbands,Guo2024SpGesture,Zabihi2023TraHGR,Salter2024emg2pose,Zhang2023LSTEMGNet,Buchner2025CVPR}.
However, sEMG signals suffer from high noise levels, large inter-subject variability, and a non-stationary temporal structure characterized by alternating transient bursts and sustained contractions~\cite{Guo2024SpGesture,Guo2025Revisiting,Lin2024RotaryTransformer,Wang2025ReactEMG}, which imposes significant challenges for accurate classification. Existing research mainly advances along two directions: one focuses on designing deeper and more expressive neural architectures to enhance signal representation~\cite{Wang2025TrustEMGNet,Buchner2025CVPR,Gu2024Mamba,Zhu2024VisionMamba,Bao2025PostStrokeDL}, while the other leverages large-scale external pretraining to introduce additional priors and improve recognition performance~\cite{Yang2023BIOT,Jin2024TimeLLM,Salter2024emg2pose}.

\begin{figure}[t]
  \centering
  \includegraphics[width=\linewidth]{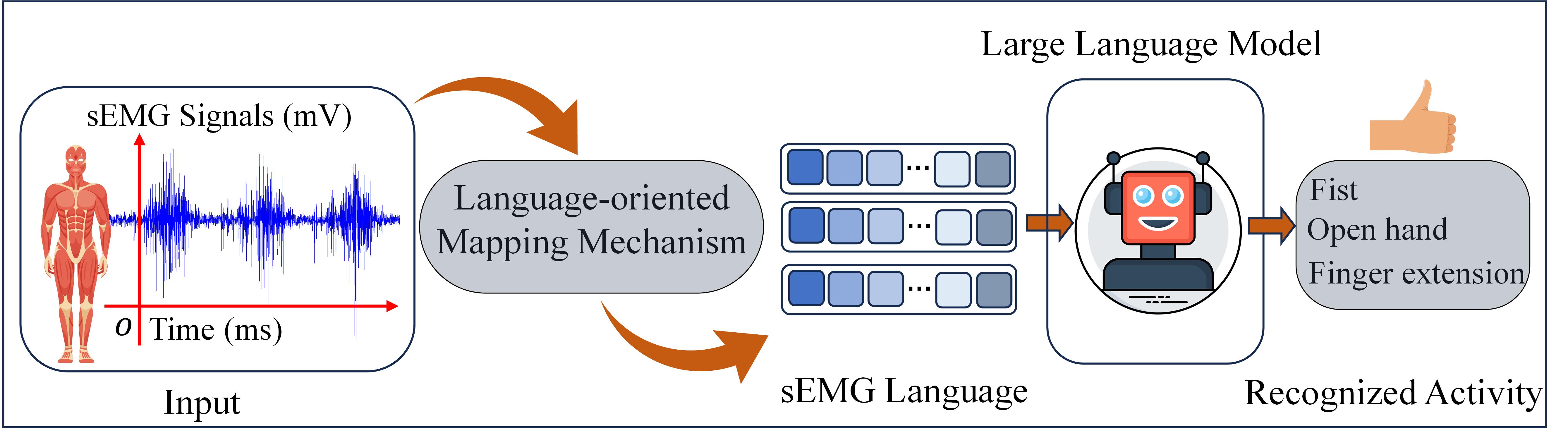}
  \caption{Illustration of the proposed LLM-sEMG framework. Continuous sEMG signals are first converted into language-like representations through a linguistic mapping process. The ``sEMG language'' is then processed by a large language model to recognize activity, with its pre-trained weights kept unchanged to retain prior knowledge.
}  
\vspace{-4mm}
  \label{fig1:framework}
\end{figure}

In recent years, large language models (LLMs) have achieved revolutionary breakthroughs in natural language processing and have gradually expanded into emerging domains such as multimodal reasoning, code comprehension, and embodied intelligence~\cite{pmlr2023,Brown2020GPT3,Li2022AlphaCode,palm2023,Liu2023Visual,Qu2024CVPR}. Recent studies~\cite{Hensel2023GestureSelection,Nyatsanga2023CoSpeechReview,Torshizi2025VirtualHumanGesture} have demonstrated that LLMs have potential applicability in sEMG-based activity recognition, as their pre-training corpora contain extensive linguistic descriptions of activities and their semantic intentions, providing implicit priors beneficial for sEMG semantic understanding. However, directly achieving activity recognition from sEMG physiological signals using LLMs still faces significant challenges. LLMs typically take natural language with sentence-level structure as input, whereas sEMG signals are continuous, non-linguistic time series that cannot be directly interpreted by LLMs. One seemingly direct approach is to fine-tune pre-trained LLMs to accept and interpret non-linguistic inputs. Indeed, recent work~\cite{Chang2025LLM4TS,palm2023,jiang2023motiongpt,Liu2023Visual} has followed this direction by fine-tuning LLMs for non-verbal tasks. Existing studies~\cite{Gururangan2020,weng2023} attempt to address this mismatch by modifying the LLM’s weights so that it can process such non-linguistic inputs. However, training an LLM in this manner on a limited sEMG dataset can lead to the loss of previously acquired knowledge, which contradicts our goal of leveraging the LLM’s rich knowledge.

By leveraging pre-training on massive natural language corpora, LLMs typically learn semantic structures and expressive patterns shared across languages. This enables them to rapidly transfer and adapt their capabilities when encountering new languages that were not covered during the training phase~\cite{Brown2020GPT3,Petroni2019,Wei2022EmergentAO,wei2022chain}. Based on the cross-linguistic generalization and aforementioned analysis, our objective is to explore how to directly apply LLMs to sEMG activity recognition while maximizing the preservation of their knowledge and semantic priors by keeping the pre-trained weights of the LLM unchanged. To this end, we propose a novel LLM-sEMG framework, as illustrated in Figure~\ref{fig1:framework}, which transforms continuous sEMG signals into a human language-like representation, referred to as the ``sEMG language.''
In the process of converting continuous sEMG signals into ``sEMG language'' that can be understood by LLMs, we propose a mechanism for sEMG-to-language emergence comprising multiple key strategies. 
Inspired by the notion that ``natural language consists of sequences of discrete lexical tokens'' and drawing on related research~\cite{Qu2024CVPR,Chang2022,Sennrich2016,Oord2017} in discrete representations, we first employ a vector quantization autoencoder (VQ-VAE) to map sEMG signals into sequences composed of discrete tokens. However, we recognize that the mere existence of a token sequence does not automatically render it suitable for LLM inference. For an LLM to accurately recognize activities based on the ``sEMG language,'' the following requirements must be met. First, since LLMs are pre-trained on vast amounts of natural language text, the ``sEMG language'' must resemble human language expressions to facilitate processing alongside natural language instructions, thereby enabling the LLM to understand it more easily. Additionally, the ``sEMG language'' must retain the discriminative information inherent in the original sEMG signals; otherwise, the LLM would naturally be unable to achieve accurate activity recognition.

To make the ``sEMG language'' closer to human natural language, we draw inspiration from the Lewis Signaling Game in linguistics~\cite{Zheng2024CVPR,ben2024,Sevestre2025,ueda2024lewiss,Rita2022Lewis}, where the encoder and decoder are regarded as two communicating agents that interact through a shared codebook. During the iterated learning process, the decoder is periodically renewed to simulate the emergence of new learners, introducing an intergenerational cultural transmission mechanism that drives the evolution of the ``sEMG language.'' However, the evolution of human language typically spans multiple generations of cultural accumulation, which is difficult to fully reproduce within limited data scales and training iterations. To address this, inspired by \cite{Qu2024CVPR,papadimitriou2023injecting}, we incorporate established human inductive biases observed in natural language~\cite{simon2007,kouwenhoven2025,Lake2023,Piantadosi2014}, such as Zipf’s law and context sensitivity, into the evolution of the ``sEMG language,'' thereby accelerating its linguistic development. To enhance its discriminative expressiveness, we also leverage reconstruction residuals to characterize local information density and perform residual-based adaptive token allocation during language emergence, effectively preserving the short-term dynamic characteristics of sEMG signals. After completing the language mapping, we employ Low-Rank Adaptation (LoRA)~\cite{hu2022lora} to adapt the LLM to interpret the ``sEMG language.'' The LLM’s pre-trained weights remain unchanged during this phase, preserving its rich prior knowledge and enabling effective activity recognition within the LLM-sEMG framework.

The main contributions are: 1) We propose a novel LLM-sEMG framework that translates sEMG signals into a human-language-like form, enabling LLMs to perform sEMG-based activity recognition without modifying their pre-trained weights. 2) We design a comprehensive sEMG-to-language mapping mechanism that integrates multiple strategies to construct a new language-like representation for sEMG signals. 3) We demonstrate through experiments on multiple public sEMG activity datasets that our method significantly improves recognition accuracy, generalization capability, and interpretability.

\section{Related Work}
\label{sec:related}

\noindent\textbf{sEMG-Based Activity Recognition.} sEMG-based activity recognition has attracted increasing research attention in recent years. sEMG provides a non-invasive means of measuring neuromuscular activity, offering a unique physiological modality for decoding human movement intentions. Early sEMG-based activity recognition methods followed classical pattern recognition frameworks centered on manually designed feature extractors~\cite{hudgins1993}. To enhance discriminative capacity, traditional machine learning models such as SVMs and HMMs were subsequently explored~\cite{Chan2005,Oskoei2008}, but deep learning has fundamentally transformed the field~\cite{allard2019,Vasanthi2024,Wei2019,Benavides2025}. Recently, Transformer-based models such as BERT and Vision Transformers (ViT) have also been applied to process sEMG signals~\cite{Guo2025Revisiting,Lin2023,Zabihi2023TraHGR,shen2022}. Furthermore, spiking neural networks (SNNs) utilize event-driven processing models to efficiently accomplish sEMG-based activity recognition tasks~\cite{Guo2024SpGesture,Kim2024,Sun2023,Tieck2020}. Despite these advances, most existing methods remain confined to signal-level modeling and fail to convert sEMG signals into human language-like representations, thereby limiting their potential to leverage the semantic reasoning capabilities of LLMs for accurate activity recognition.

\noindent\textbf{Large Language Models.} LLMs such as GPT~\cite{OpenAI2023GPT4}, LLaMA~\cite{touvron2023llama}, and Mistral~\cite{Jiang2023Mistral7B} have made significant strides in natural language processing tasks. Through pre-training on web-scale multimodal corpora encompassing trillions of tokens, they have achieved unprecedented scale~\cite{Brown2020GPT3,Chowdhery2023PaLM,Fedus2022,Jiang2024Mixtral}. This endows them with rich implicit knowledge and semantic priors, enabling exceptional capabilities in complex reasoning and abstract pattern recognition~\cite{Kojima2022,Li2022AlphaCode,xie2022,Petroni2019,Mirchandani2023,Wei2022EmergentAO}. Recent research~\cite{Gemini2024,Qu2024CVPR,Lewkowycz2022,Radford2021CLIP,wei2022chain} indicates that LLMs can also be extended to non-linguistic domains such as multimodal understanding and structured reasoning, demonstrating immense potential as universal knowledge engines. 
\section{Proposed Method}
\label{sec:proposed}

\begin{figure*}[t]
  \centering
  \includegraphics[width=\textwidth]{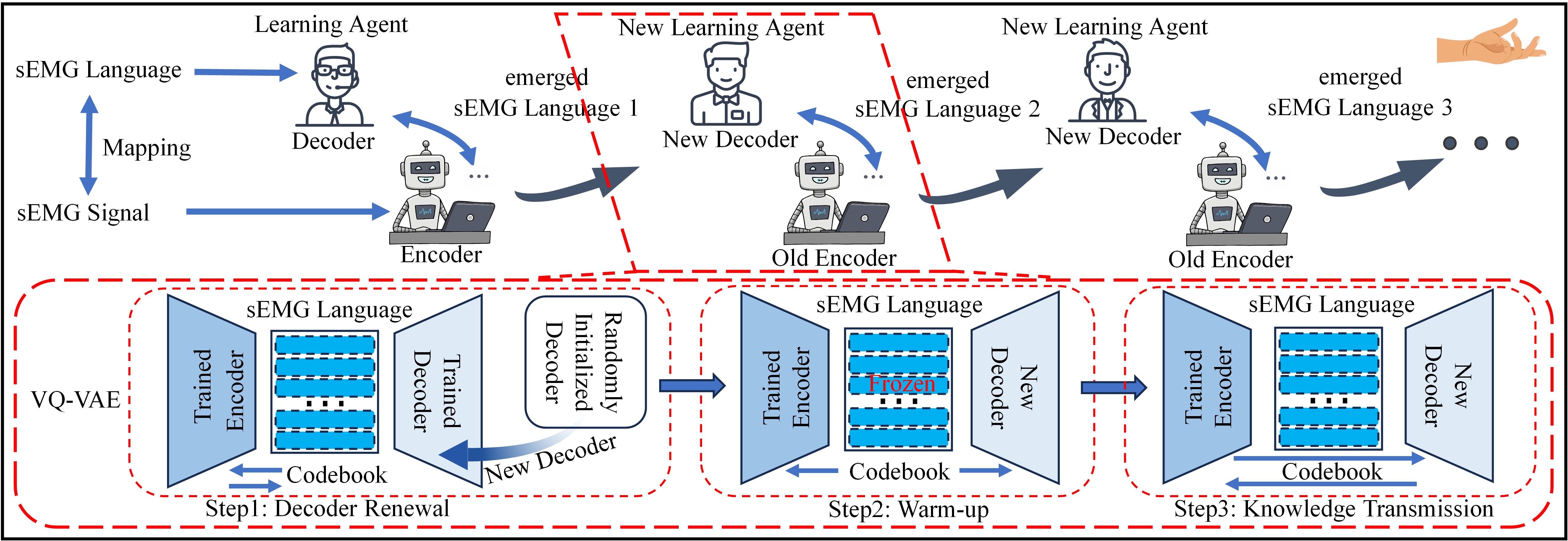}
  \caption{Illustration of the proposed framework for sEMG-to-language emergence. Each generation renews the decoder while freezing the codebook to ensure stable knowledge transfer. The process includes three steps: decoder renewal, warm-up, and knowledge transmission.}
  \label{fig2:sEMG-to-Language}
\end{figure*}

This section details our proposed LLM-sEMG framework, which leverages LLMs for accurate recognition of sEMG activity. However, a representational gap exists between the non-linguistic nature of sEMG signals and the linguistic input format required by LLMs. To address this, the framework converts sEMG signals into human-language-like representations, referred to as ``sEMG language,'' which are then fed into an LLM to infer the corresponding activities. The subsequent sections detail the language emergence process, the feature preservation mechanism, and the training and inference phases.

\subsection{sEMG-to-Language Emergence}


Inspired by the linguistic principle that natural language sentences are composed of discrete symbolic units~\cite{Sennrich2016,Mielke2021}, and the success of VQ-VAE in discretizing continuous signals into discrete token sequences~\cite{Oord2017},
we propose an sEMG-tailored VQ-VAE model to discretize continuous signals into tokens. Building on this foundation, and inspired by~\cite{Zheng2024CVPR}, we adopt the Lewis Signaling Game framework~\cite{Lewis2002,Rita2022Lewis} to simulate human linguistic and cultural transmission mechanisms, thereby enabling the emergence of an ``sEMG language'' that can be processed by LLMs. Furthermore, motivated by~\cite{Qu2024CVPR,papadimitriou2023injecting}, we incorporate human inductive biases into the language evolution process to accelerate the convergence of the generated ``sEMG language'' toward human natural language.

\textbf{sEMG-Oriented VQ-VAE Model.}
The overall architecture of the discretization model shares similarities with previous VQ-VAE designs~\cite{Qu2024CVPR,Zhang2023regul,jiarong2023} and comprises three core components: an encoder $E$, a codebook $\mathcal{C}$, and a decoder $D$. The encoder is responsible for feature extraction; the quantization module selects the closest prototype vector from the codebook to generate discrete tokens; and the decoder attempts to reconstruct the input signal based on these tokens, thereby achieving a mapping from the continuous signal space to the linguistic space. 


Given an sEMG signal $\mathbf{X} \in \mathbb{R}^{T \times C}$, where $T$ is the total number of time steps and $C$ is the number of electromyography acquisition channels, we first divide it into overlapping segments $\mathbf{x} \in \mathbb{R}^{t \times C}$ of length $t$ using a fixed-length sliding window. The encoder $E$ then maps each segment $\mathbf{x}$ to a sequence of continuous latent feature vectors $\mathbf{z}_e(\mathbf{x}) = E(\mathbf{x}) = 
\begin{bmatrix} 
\mathbf{z}_{e,1} & \mathbf{z}_{e,2} & \cdots & \mathbf{z}_{e,S} 
\end{bmatrix} \in \mathbb{R}^{D \times S}$, 
where $S$ is the length of the latent feature sequence and $\mathbf{z}_{e,s} \in \mathbb{R}^{D}$. Each latent vector $\mathbf{z}_{e,s}$ is obtained via a one-dimensional convolutional neural network (1D-CNN) to capture the short-term dynamics of the sEMG signal.

To discretize the sequence of continuous latent feature vectors $\mathbf{z}_e(\mathbf{x})$, we construct a codebook consisting of $K$ learnable vectors $\mathcal{C} = \{\mathbf{e}_j\}_{j=1}^{K}$, where $ \mathcal{C} \in \mathbb{R}^{D \times K}$, $K$ is the total number of tokens in the codebook, and $D$ is the dimensionality of each token vector, matching the token dimension expected by the subsequent language model. For each position $s$ in the latent feature sequence, the closest codebook vector is selected by computing the Euclidean distance, yielding the corresponding discrete token. This selection is formalized as
$
k_s = \arg\min_{j \in \{1, \dots, K\}} \left\| \mathbf{z}_{e,s} - \mathbf{e}_j \right\|_2
$,
where $\mathbf{z}_{e,s} \in \mathbb{R}^D$ is the $s$-th latent vector from the encoder and $\mathbf{e}_j$ is the $j$-th codebook vector. 
The discrete latent vector corresponding to position $s$ is then defined as $\mathbf{z}_{q,s}(\mathbf{x}) = \mathbf{e}_{k_s}.$
Consequently, the input segment $\mathbf{x}$ is mapped to a discrete token sequence
$\mathbf{z}_q(\mathbf{x}) = [\mathbf{z}_{q,1}, \mathbf{z}_{q,2}, \dots, \mathbf{z}_{q,S}].$

Finally, the decoder $D$ reconstructs the input segment from the discrete sequence, generating the predicted output $\hat{\mathbf{x}}$. Through this process of encoding, codebook quantization, and decoding, the continuous sEMG signal is mapped into a discrete token sequence for efficient signal-to-language conversion.


\textbf{Language Emergence Strategy.} Although VQ-VAE can convert continuous sEMG signals into discrete token sequences, these sequences do not inherently resemble human natural language. This lack of linguistic properties prevents LLMs from activating their latent knowledge of sEMG semantics and the corresponding activities, thereby limiting their reasoning capabilities. Inspired by the ``cultural transmission theory'' in language evolution~\cite{Zheng2024CVPR,Carr2017Struc,Smith2008Cul,simon2007}, we introduce the Lewis Signaling Game framework~\cite{Lewis2002,Rita2022Lewis} to simulate the emergence mechanism of human language through successive generations of cultural transmission. The Lewis Signaling Game describes a communication process between two participants who initially interact through symbols that are mutually ``unintelligible.'' Through repeated communication, a shared and interpretable language gradually emerges between them. Inspired by this concept, we reinterpret the encoder and decoder in the VQ-VAE model as the two participants in a Lewis Signaling Game, where the codebook serves as the shared set of communication symbols. By periodically replacing the decoder within the VQ-VAE, we simulate the emergence of new ``generations,'' enabling the ``sEMG language'' to evolve and emerge across intergenerational transmission. As illustrated in Figure~\ref{fig2:sEMG-to-Language}, during the emergence of the ``sEMG language,'' each generation corresponds to a process of language learning and transmission, consisting of the following three key steps.

(1) Decoder Renewal: During the iterated learning process, we simulate the emergence of new learners in cultural transmission by introducing a ``new learning agent'' in each generation. Specifically, the encoder consistently assumes the role of a ``speaker'' during iterative learning. It must continuously learn across generations how to map raw sEMG signals into semantically distinct discrete tokens, gradually establishing a stable and transferable semantic mapping between the sEMG signals and the ``sEMG language.'' In contrast, the decoder's task more closely resembles that of a “new generation learner” in cultural transmission. It does not redefine the linguistic system but instead learns from the existing codebook how to interpret and reconstruct ``sEMG language'', thereby progressively mastering and perpetuating the ``sEMG language'' formed by the preceding generation. Therefore, at the start of each iteration, we replace the decoder within the VQ-VAE model, introducing it as a newly initialized ``learning agent.'' This process mirrors the scenario described in cultural transmission theory, where language must be relearned by the next generation of learners, thereby simulating the beginning of a new learning cycle. Moreover, in the new generation, the encoder and codebook parameters are preserved from the preceding generation and remain fixed. This allows the encoder and codebook to serve as “knowledge carriers” representing the language evolution process and ensures their continuity across generations. Consequently, the ``sEMG language'' gradually evolves to resemble human natural language through continuous iterative learning. This process can be abstracted as follows: $ E^{(t+1)} = E^{(t)}, \quad \mathcal{C}^{(t+1)} = \mathcal{C}^{(t)}, \quad D^{(t+1)} \sim \mathcal{P}_{\text{init}}$, where $E^{(t)}$ and $\mathcal{C}^{(t)}$ denote the encoder and the codebook after completing training at generation $t$, respectively, while $\mathcal{P}_{\text{init}}$ represents the initial parameter distribution of the newly initialized agent.

(2) Warm-up: Since the newly introduced learning agent starts from random initialization, it has not yet acquired the semantic mapping relationship formed between the encoder and the codebook in the previous generation. Joint training with the encoder and codebook at this step could induce substantial fluctuations in the codebook gradients, potentially destabilizing the ``sEMG language'' established in the previous generation.  To prevent such degradation, we introduce a brief warm-up stabilization step before initiating the new generation of training. In this step, the codebook $\mathcal{C}^{(t)}$ is frozen, and only the newly initialized decoder $D^{(t+1)}$ is trained for a few iterations to quickly adapt to the frozen codebook. This short adaptation allows the new decoder to inherit the ``sEMG language'' established in the previous generation without introducing disruptive gradient updates. 

(3) Knowledge Transmission:
After the new learning agent completes its initial adaptation to the codebook, we unfreeze the codebook and continue training the model following the reconstruction learning paradigm of VQ-VAE. The objective of this stage is to achieve cross-generational knowledge transfer through the shared codebook, enabling the new learning agent to exchange knowledge with the encoder. This process drives the continuous evolution of the ``sEMG language,'' gradually endowing it with representational characteristics more closely aligned with human natural language. To induce a learning bottleneck, we deliberately maintain an incomplete and biased “teaching” process from the old encoder to the new decoder, thereby continuously stimulating the spontaneous evolution of the language. Upon completion of the knowledge transfer phase, training for the current generation is complete, and the training for the next generation of learning agents begins. 

The above three steps are executed cyclically until training concludes. In the final generation, we extend the knowledge-transfer step to ensure convergence and the formation of a stable ``sEMG language.''
 
\textbf{Human-Linguistic Bias Guided Learning.}
The evolution of natural language is a protracted process. Words and sentences do not emerge randomly but gradually evolve through long-term intergenerational transmission into forms that are easier for humans to comprehend, remember, and utilize~\cite{Arnon2024Cul,Tamariz2016Cul}. Previous research~\cite{ben2024,simon2007,kouwenhoven2025,Piantadosi2014} has demonstrated that this evolutionary process is influenced by a series of human linguistic biases, and that humans tend to follow certain prior distributions reflecting these preferences. Therefore, incorporating such linguistic biases into the iterative learning framework can further accelerate the evolution of ``sEMG language'' and make it more suitable for subsequent processing by large language models. During iterative learning, expression patterns that are more easily understood and reused by new decoders are progressively retained and amplified through training~\cite{Zheng2024CVPR}. This tendency gives rise to a word frequency structure in which a small number of high-frequency core tokens coexist with a large number of low-frequency extension tokens~\cite{he2025pre}. Furthermore, as the iterative process progresses, the encoder gradually favors generating sequences with contextual dependencies to enhance semantic comprehension for successive generations~\cite{Rita2022Lewis,Qu2024CVPR}. Motivated by these observations, we explicitly model two key human linguistic biases, namely Zipf’s law and context sensitivity, to accelerate the evolution of ``sEMG language.''

(1) Zipf’s Law Prior:
Zipf’s law describes the power-law distribution of word frequencies in natural languages~\cite{he2025pre,LaviRotbain2022Zif}. To simulate this linguistic regularity in ``sEMG language,'' we introduce this bias during the iterative evolution process. However, since the model has not yet established stable token usage patterns in the early stages, directly introducing the Zipf regularization term may cause instability in the codebook gradients. Therefore, during the initial iterations (approximately 25\% of the total training process), we employ Zipf-weighted sampling to make a small subset of “core word tokens” more likely to be activated. This initialization naturally induces a “frequent–rare word” distribution similar to that of human language, providing the newly generated codebook $\mathcal{C}$ with a Zipf-law prior. As a result, the evolution of ``sEMG language'' is accelerated toward linguistically coherent organizational patterns from the very beginning of training. In subsequent iterative learning, we further introduce a Zipf regularization term to ensure that the ``sEMG language'' maintains a stable Zipf bias throughout the iterative evolution process. This encourages the overall token usage frequency distribution to conform to Zipf’s law. Specifically, we compute the empirical distribution $D_{\mathrm{freq}}$ of token usage in the current batch and align it with the theoretical Zipf distribution $D_{\mathrm{Zipf}}$ using the Jensen-Shannon divergence, formulated as $\mathcal{L}_{\text{zipf}} = \text{JS}\left(D_{\text{freq}} \parallel D_{\text{Zipf}}\right)$, where $\text{JS}(\cdot\parallel\cdot)$ denotes the Jensen-Shannon divergence, which quantifies the difference between two probability distributions.

(2) Context-Sensitivity Prior:
Beyond word frequency imbalance, human natural language exhibits significant context sensitivity~\cite{Garten2019Context,Yoon2025Lex}. In other words, the usage of word tokens is often not independent but depends on their contextual relationships, revealing distinct semantic associations. This context sensitivity serves as a crucial foundation for achieving language comprehension and expression. To incorporate similar contextual properties into the emergent ``sEMG language,'' we involve a mechanism that integrates both intra-generational formation and inter-generational preservation of contextual dependencies. This enables the ``sEMG language'' to rapidly develop contextual relationships within each generation while maintaining stable dependency patterns across successive generations during knowledge transmission.

Specifically, during each generation of evolution, we encourage semantically related word tokens within the ``sEMG language'' to co-occur in the same sequence, thereby gradually forming a context-sensitivity bias similar to that observed in human language. To achieve this, we introduce a context-sensitivity loss function, defined as
$
\mathcal{L}_{\text{context}} = 1 - \frac{1}{N}\sum_{n=1}^{N} \text{Corr}(t_n)
$,
where $N$ denotes the batch size, and $\text{Corr}(t_n)$ quantifies the degree of correlation among semantically related tokens within the $n$-th sequence. The exact definition of $Corr(\cdot)$ is provided in the supplementary material. By minimizing $\mathcal{L}_{\text{context}}$, the model is encouraged to generate more semantically correlated tokens within each generation, enabling the ``sEMG language'' to gradually exhibit the local semantic continuity characteristic of human natural language. Once such contextual dependencies are gradually formed intra-generationally, we further aim to ensure their stable inheritance inter-generationally. Specifically, to ensure that the inter-generational context-sensitivity bias is consistently preserved, we introduce a bias-preserving regularization term based on the token co-occurrence matrix $M^{(t)}$, formulated as 
$\mathcal{L}_{\text{preserving}} = \left\| M^{(t+1)} - M^{(t)} \right\|_F^2$, 
where $\|\cdot\|_F$ denotes the Frobenius norm. The detailed computation of $M^{(t)}$ is provided in the supplementary material. This regularization term ensures that the context-sensitivity bias of the ``sEMG language'' remains consistent across inter-generational transitions, allowing the ``sEMG language'' to gradually inherit this bias during evolution. 

Based on the aforementioned human linguistic biases, we define the human-linguistic-bias-guided learning objective as follows:
\begin{equation}
\mathcal{L}_{\text{human}} = \mathcal{L}_{\text{zipf}} + 
\mathcal{L}_{\text{context}} + 
\mathcal{L}_{\text{preserving}}
\label{eq:human_loss}
\end{equation}

Through joint optimization, the ``sEMG language'' gradually evolves linguistic regularities during iterative learning, making it more interpretable when used as input to large language models.

\subsection{sEMG Feature-Preserving Tokenization}

The core challenge of sEMG signals lies in their highly dynamic and non-stationary nature, as muscle activation patterns associated with different activities exhibit substantial temporal variations in information density~\cite{SUN2020Intelli,Chowdhury2013SurfaceEMG}. Rapid, high-amplitude bursts typically indicate the initiation, peak force, or termination of a movement~\cite{Liu2015EBPP}, providing rich discriminative information crucial for activity recognition~\cite{Crotty2021Onset}. In contrast, steady-state activities correspond to sustained contractions or rest phases, where sEMG signals are relatively stable and contribute less novel information~\cite{Liu2021WRHand,Guo2024SpGesture}. Under these characteristics, the standard uniform tokenization strategy allocates tokens over fixed temporal intervals, failing to capture the varying information density. In information-rich transient phases, too few tokens are assigned, while in steady phases, too many tokens introduce redundancy, which is unfavorable for LLM reasoning. To address this, we adaptively allocate a fixed token budget, concentrating representational capacity on the most discriminative segments. Prior studies~\cite{Oh2018Residual,Xu2022Anomaly,cai2025self,Audibert2020USAD} have shown that reconstruction residuals effectively reflect the local information density of the original signal. Based on this observation, we assign more tokens to regions with higher residual energy, enabling the emergent ``sEMG language'' to better represent the key dynamic variations of activities. Conversely, stable regions undergo sparse tokenization. This adaptive allocation mechanism allows the emergent ``sEMG language'' to preserve the inherently non-stationary and burst-like nature of sEMG signals, while simultaneously reducing redundant representations during steady-state phases.

To achieve the aforementioned objectives, we represent each sEMG segment $\mathbf{x}$ as a sequence of $S$ temporal slices, $\mathbf{x}={\mathbf{x}_1,\dots,\mathbf{x}_S}$. Given the $s$-th slice and its reconstruction $\hat{\mathbf{x}}_{s}$ from the VQ-VAE decoder, the residual energy is defined as
$ R_{s} = \big\| \mathbf{x}_{s} - \hat{\mathbf{x}}_{s} \big\|_{2}^{2}$.
High residuals indicate non-stationary or abrupt changes, whereas low residuals correspond to stable regions. The residuals are normalized into a probability distribution $P_s$, which provides an importance weighting for token allocation. Given a fixed token budget $T_{\text{max}}$, the initial allocation is $T_s = T_{\text{max}} \times P_s$. 
To avoid losing stable regions, a minimum-coverage constraint ensures that every slice receives at least one token. If the total number of assigned tokens exceeds $T_{\text{max}}$, the surplus tokens are subtracted from slices with the smallest residuals. 
Under this constraint, more tokens are allocated to high-residual slices and fewer to stable ones. 
Detailed computational formulas are provided in the Supplementary Material.

\subsection{Training and Testing}
This section presents the training and testing scheme of the LLM-sEMG framework, which integrates linguistic emergence, adaptive tokenization, and LLM-based reasoning for activity recognition. The process consists of two stages: (1) learning an iteratively evolving, feature-preserving VQ-VAE that maps continuous sEMG signals into ``sEMG language,'' and (2) fine-tuning an LLM via LoRA~\cite{hu2022lora} to recognize activities from the ``sEMG language.''

\textbf{Training.} In the first stage, we train an sEMG-oriented VQ-VAE that jointly performs language emergence and adaptive tokenization under an iterated-learning framework.
Given a batch of input sEMG sequences
$\{\mathbf{x}^b\}_{b=1}^{B}$, where $\mathbf{x}^b \in \mathbb{R}^{T \times C}$, $b$ indexes the samples in the batch, and $B$ denotes the batch size. The encoder $E$ extracts a sequence of continuous latent feature vectors for each input $\mathbf{x}^b$, i.e., $\mathbf{z}_e(\mathbf{x^b}) = E(\mathbf{x^b}) = [\mathbf{z^b}_{e,1}, \dots, \mathbf{z^b}_{e,S}] \in \mathbb{R}^{D \times S}$. These latent vectors are quantized by the learnable codebook $\mathcal{C}$ to form the discrete ``sEMG language'' representation $\mathbf{z}_{q}(\mathbf{x}^b)$. Finally, the decoder $D$ reconstructs the input $\hat{\mathbf{x}}^b$ from the quantized representation $\mathbf{z}_{q}(\mathbf{x}^b)$.
Following previous VQ-VAE studies~\cite{Oord2017,Qu2024CVPR}, three typical loss terms are jointly optimized, namely the reconstruction loss $\mathcal{L}_{\text{rec}}$, the embedding loss $\mathcal{L}_{\text{emb}}$, and the commitment loss $\mathcal{L}_{\text{com}}$, formulated as follows:
\begin{equation}
\begin{aligned}
\mathcal{L}_{\text{rec}} &=\text{Smooth-}L_1(\{\mathbf{x}^b\}_{b=1}^{B},\{\hat{\mathbf{x}}^b\}_{b=1}^{B}) \\
\mathcal{L}_{\text{emb}} &= \|\text{sg}(\mathbf{z}_e(\{\mathbf{x}^b\}_{b=1}^{B})) - \mathbf{z}_q(\{\mathbf{x}^b\}_{b=1}^{B})\|_2^2 \\
\mathcal{L}_{\text{com}} &= \|\mathbf{z}_e(\{\mathbf{x}^b\}_{b=1}^{B}) - \text{sg}(\mathbf{z}_q(\{\mathbf{x}^b\}_{b=1}^{B}))\|_2^2.
\end{aligned}
\label{eq:vqvae_basic_losses}
\end{equation}

To make the learned representation linguistically meaningful, we introduce an iterated learning mechanism. The encoder, decoder, and codebook are regarded as the three parties involved in the Lewis Signaling Game. Across generations, the encoder and codebook ($E^{(t)}, \mathcal{C}^{(t)}$) are retained as “knowledge carriers,” while the decoder ($D^{(t)}$) is periodically replaced by a newly initialized learning agent $D^{(t+1)}$. During each generation, the encoder performs adaptive token allocation by dynamically determining the number of tokens assigned to each temporal slice based on its local reconstruction residual. Temporal slices with higher residuals are allocated more tokens to emphasize transient and information-rich dynamics, whereas stable regions are sparsely tokenized to reduce redundancy. Before resuming joint training, a warm-up step is conducted to align the newly initialized decoder with the frozen codebook, enabling smooth knowledge transfer and stable language evolution. To further accelerate linguistic convergence, we incorporate human-linguistic-bias guidance, including Zipf’s law and context-sensitivity regularization, as defined in Eq.~(\ref{eq:human_loss}). The total training loss for this stage is therefore:
\begin{equation}
\mathcal{L}_{\text{total}} =
\mathcal{L}_{\text{rec}} +
\mathcal{L}_{\text{emb}} +
\lambda_1 \mathcal{L}_{\text{com}} +
\lambda_2 \mathcal{L}_{\text{human}}
\label{eq:total_vqvae_loss}
\end{equation}
where $\lambda_1$ and $\lambda_2$ control the relative importance of commitment and linguistic-bias regularization.

In the second stage, we fine-tune a pre-trained LLM using Low-Rank Adaptation (LoRA) to enable activity understanding and classification based on the generated ``sEMG language'' while preserving the original pre-trained weights. This adaptation allows the LLM to understand and classify the ``sEMG language'' generated by the VQ-VAE in Stage 1 without altering its intrinsic latent knowledge of sEMG semantics and the corresponding activities.
Each training sample consists of an sEMG token sequence and its corresponding activity label. We formulate the instruction prompt as: ``\textit{Given a sequence of sEMG tokens [tokens], please predict the corresponding activity.}'' where [tokens] represents the ``sEMG language'' sequence generated in the first stage. During LoRA fine-tuning, the model achieves this objective by minimizing the cross-entropy loss between the predicted activity token $t_p$ and the ground-truth label $t_g$, i.e.,
$
\mathcal{L}_{\text{LLM}} = \mathcal{L}_{\text{ce}}(t_p, t_g).
$
This enables the LLM to perform efficient and accurate activity recognition based on the ``sEMG language.''

\textbf{Inference and Testing.} During the testing phase, each sEMG signal is first processed by the trained VQ-VAE for adaptive discretization, generating the corresponding ``sEMG language'' sequence. The resulting sequence is then fed into the LoRA-fine-tuned LLM, which predicts the activity category using the same instruction format as in the training stage. Together, these two stages establish an end-to-end reasoning from continuous physiological sEMG signals to activities, effectively enabling the large language model to function as an sEMG-based activity recognizer.

\section{Experiments}
\label{sec:experiments}

\begin{table}[t]
\centering
\caption{Comparison of sEMG-based activity recognition methods on the GRABMyo dataset in terms of ACC (\%) and STD.}
\vspace{-3mm}
\label{tab:gesture_results1}
\setlength{\tabcolsep}{4pt}
\small
\resizebox{\columnwidth}{!}{%
\begin{tabular}{lcccccccccc}
\toprule
\multirow{2}{*}{\centering\textbf{Model}} &
\multicolumn{2}{c}{\textbf{Single-finger}} &
\multicolumn{2}{c}{\textbf{Multi-finger}} &
\multicolumn{2}{c}{\textbf{Wrist}} &
\multicolumn{2}{c}{\textbf{Rest}} &
\multicolumn{2}{c}{\textbf{Overall}} \\
\cmidrule(lr){2-3} \cmidrule(lr){4-5} \cmidrule(lr){6-7} \cmidrule(lr){8-9} \cmidrule(lr){10-11}
& \textbf{ACC} & \textbf{STD}
& \textbf{ACC} & \textbf{STD}
& \textbf{ACC} & \textbf{STD}
& \textbf{ACC} & \textbf{STD}
& \textbf{ACC} & \textbf{STD} \\
\midrule
TCN~\cite{Tsinganos2019TCN}        & 78.78 & 0.02 & 79.10 & 0.02 & 87.27 & 0.01 & 88.57 & 0.02 & 81.50 & 0.02 \\
Asif~\cite{Asif2020}            & 83.44 & 0.02 & 83.58 & 0.01 & 89.40 & 0.01 & 90.86 & 0.01 & 85.34 & 0.01 \\
GRU~\cite{Chen2021GRU}             & 84.45 & 0.02 & 84.88 & 0.01 & 90.06 & 0.01 & 89.42 & 0.02 & 86.30 & 0.02 \\
Zerveas~\cite{zerveas2021}      & 78.45 & 0.02 & 77.20 & 0.02 & 87.28 & 0.02 & 86.76 & 0.02 & 80.43 & 0.02 \\
Informer~\cite{Zhou2021Informer}        & 86.88 & 0.02 & 86.54 & 0.02 & 91.90 & 0.01 & 83.56 & 0.02 & 87.71 & 0.02 \\
TEMGNet~\cite{rahimian2021temg}       & 77.70 & 0.02 & 74.00 & 0.03 & 84.04 & 0.02 & 87.46 & 0.01 & 78.02 & 0.02 \\
LST-EMG-Net~\cite{Zhang2023LSTEMGNet}    & 87.21 & 0.01 & 83.16 & 0.01 & 88.36 & 0.02 & 82.52 & 0.02 & 85.31 & 0.02 \\
JASNN~\cite{Guo2024SpGesture}            & 85.55 & 0.01 & 86.62 & 0.01 & 87.82 & 0.01 & 86.55 & 0.01 & 86.05 & 0.01 \\
BeyondVision~\cite{Wang2024Beyond}    & 81.73 & 0.02 & 85.92 & 0.02 & 87.48 & 0.02 & 87.79 & 0.02 & 85.82 & 0.02 \\
STET~\cite{Guo2025Revisiting}             & 88.27 & 0.01 & 89.93 & 0.02 & 93.77 & 0.01 & 95.33 & 0.01 & 90.76 & 0.01 \\
\textbf{LLM-sEMG (Ours)}        & \textbf{92.25} & 0.02 & \textbf{94.50} & 0.01 & \textbf{95.19} & 0.01 & \textbf{96.82} & 0.01 & \textbf{95.14} & 0.01 \\
\bottomrule
\end{tabular}%
}
\end{table}

\begin{table}[t]
\centering
\caption{Comparison of sEMG-based activity recognition methods on the NinaPro DB2 dataset in terms of ACC (\%) and STD.}
\vspace{-3mm}
\label{tab:gesture_results2}
\setlength{\tabcolsep}{4pt}
\small
\resizebox{\columnwidth}{!}{%
\begin{tabular}{lcccccccccc}
\toprule
\multirow{2}{*}{\centering\textbf{Model}} &
\multicolumn{2}{c}{\textbf{Exercise A}} &
\multicolumn{2}{c}{\textbf{Exercise B}} &
\multicolumn{2}{c}{\textbf{Exercise C}} &
\multicolumn{2}{c}{\textbf{Rest}} &
\multicolumn{2}{c}{\textbf{Overall}} \\
\cmidrule(lr){2-3} \cmidrule(lr){4-5} \cmidrule(lr){6-7} \cmidrule(lr){8-9} \cmidrule(lr){10-11}
& \textbf{ACC} & \textbf{STD}
& \textbf{ACC} & \textbf{STD}
& \textbf{ACC} & \textbf{STD}
& \textbf{ACC} & \textbf{STD}
& \textbf{ACC} & \textbf{STD} \\
\midrule
TCN~\cite{Tsinganos2019TCN}        & 74.45 & 0.02 & 74.67 & 0.03 & 79.30 & 0.02 & 76.52 & 0.02 & 75.23 & 0.02 \\
Asif~\cite{Asif2020}            & 76.12 & 0.02 & 76.62 & 0.02 & 80.17 & 0.02 & 82.49 & 0.01 & 78.55 & 0.02 \\
GRU~\cite{Chen2021GRU}             & 78.50 & 0.02 & 78.33 & 0.02 & 81.41 & 0.02 & 81.95 & 0.02 & 80.17 & 0.02 \\
Zerveas~\cite{zerveas2021}      & 81.32 & 0.03 & 78.50 & 0.02 & 83.29 & 0.02 & 84.07 & 0.02 & 81.75 & 0.02 \\
Informer~\cite{Zhou2021Informer}        & 82.64 & 0.02 & 83.84 & 0.02 & 85.22 & 0.01 & 85.12 & 0.01 & 84.05 & 0.01 \\
TEMGNet~\cite{rahimian2021temg}       & 78.27 & 0.02 & 78.05 & 0.03 & 80.44 & 0.02 & 82.42 & 0.02 & 79.59 & 0.02 \\
LST-EMG-Net~\cite{Zhang2023LSTEMGNet}    & 85.31 & 0.02 & 85.74 & 0.01 & 87.23 & 0.02 & 85.61 & 0.01 & 85.72 & 0.01 \\
JASNN~\cite{Guo2024SpGesture}            & 85.73 & 0.02 & 85.12 & 0.02 & 87.25 & 0.02 & 85.90 & 0.01 & 85.80 & 0.02 \\
BeyondVision~\cite{Wang2024Beyond}    & 85.60 & 0.02 & 84.38 & 0.03 & 86.71 & 0.02 & 86.84 & 0.01 & 85.52 & 0.02 \\
STET~\cite{Guo2025Revisiting}             & 88.65 & 0.02 & 87.22 & 0.01 & 89.35 & 0.02 & 91.32 & 0.01 & 89.13 & 0.02 \\
\textbf{LLM-sEMG (Ours)}        & \textbf{93.42} & 0.02 & \textbf{92.53} & 0.01 & \textbf{93.80} & 0.01 & \textbf{94.56} & 0.01 & \textbf{93.17} & 0.01 \\
\bottomrule
\end{tabular}%
}
\vspace{-2mm}
\end{table}

\subsection{Datasets}
To comprehensively evaluate the proposed LLM-sEMG framework, we conduct experiments on two widely used public datasets: GRABMyo~\cite{GRAB2022} and NinaPro DB2~\cite{Atzori2014Ninapro}. Following prior work, including the recent strong baseline STET~\cite{Guo2025Revisiting}, all experiments adopt the user-specific evaluation protocol. \textbf{GRABMyo}~\cite{GRAB2022} contains over 250,000 multi-channel sEMG sequences from multiple subjects performing 17 activities, including 16 dynamic activities and one resting state. \textbf{NinaPro DB2}~\cite{Atzori2014Ninapro} includes recordings from 40 subjects performing 50 hand and wrist activities captured by 12 electrodes. These datasets enable a systematic evaluation of the proposed LLM-sEMG framework under diverse signal conditions.

\subsection{Experimental Setup}
\label{sec:experimental_setup}
\textbf{Implementation Details.}
All experiments are conducted on NVIDIA H100 GPUs using the PyTorch framework, with LLaMA-13B employed as the large language model via Hugging Face. For the sEMG-oriented VQ-VAE training, the codebook size is set to $K=512$, and the token dimension is aligned with the word embedding dimension of LLaMA-13B ($D=5120$). The model is trained using the AdamW optimizer with an initial learning rate of $1\times10^{-4}$ and a batch size of 320. During iterated learning, the decoder is periodically reinitialized to simulate new generations, with each iteration beginning with a 200-step warm-up followed by an 8000-step knowledge-transfer stage; the final generation is trained for 1.5k steps. For LLM fine-tuning, we build upon the open-source implementation~\cite{litllama2023} and apply Low-Rank Adaptation (LoRA) to the pre-trained LLaMA-13B using a batch size of 320 and a learning rate of $2\times10^{-5}$.
Additional implementation and training details are provided in the Supplementary Material.

\subsection{Comparison with Baselines}
We evaluated the accuracy (ACC) and standard deviation (STD) of the proposed LLM-sEMG model on the GRABMyo and NinaPro DB2 datasets, comparing it with a range of representative sEMG-based activity recognition methods. Specifically, we compare our model with temporal convolutional networks (TCN~\cite{Tsinganos2019TCN}), recurrent architectures (GRU~\cite{Chen2021GRU}), and Transformer-based methods (Informer~\cite{Zhou2021Informer}, BeyondVision~\cite{Wang2024Beyond}, STET~\cite{Guo2025Revisiting}). Tables~\ref{tab:gesture_results1} and~\ref{tab:gesture_results2} summarize the quantitative results on the GRABMyo and NinaPro DB2 datasets, respectively.
As shown in Table~\ref{tab:gesture_results1}, our method achieves an overall accuracy of 95.14\% on the GRABMyo dataset, outperforming the previous best method STET~\cite{Guo2025Revisiting} by 4.38\%. Across all activity categories, LLM-sEMG consistently delivers the highest performance, reaching 92.25\% for single-finger, 94.50\% for multi-finger, 95.19\% for wrist, and 96.82\% for rest activities. These results demonstrate the effectiveness of the proposed framework.
On the NinaPro~DB2 dataset, as shown in Table~\ref{tab:gesture_results2}, LLM-sEMG achieves an overall accuracy of 93.17\%, surpassing the strongest baseline STET~\cite{Guo2025Revisiting} by 4.04\%. The improvements are consistent across all subsets, with gains of 4.77\% on Exercise~A, 5.31\% on Exercise~B, 4.45\% on Exercise~C, and 3.24\% on Rest. These results further validate the effectiveness of LLM-sEMG.

\subsection{Ablation Studies}
To comprehensively evaluate the contribution of each component in the LLM-sEMG framework, we conduct detailed ablation experiments on the GRABMyo dataset. 

\noindent\textbf{Effect of Iterated Learning.} 
We first analyze the impact of the iterated learning mechanism. As shown in Table~\ref{tab:ablation_iter}, removing the iterated learning mechanism (``w/o Iteration'') reduces the recognition accuracy from 95.14\% to 79.59\%, indicating that intergenerational renewal is essential for stably forming an ``sEMG language'' that can be effectively understood and processed by the LLM. In addition, removing the warm-up phase (``w/o Warm-up'') decreases the accuracy to 91.65\% and causes oscillations in the reconstruction loss during the early stage of training, demonstrating that the warm-up step plays a crucial role in ensuring smooth cross-generational knowledge transfer and maintaining semantic continuity.

\begin{table}[t]
\centering
\vspace{-3mm}
\caption{Effect of iterated learning in terms of ACC (\%) and STD.}
\vspace{-4mm}
\label{tab:ablation_iter}
\setlength{\tabcolsep}{4pt}
\small
\resizebox{\columnwidth}{!}{%
\begin{tabular}{lcccccccccc}
\toprule
\multirow{2}{*}{\centering\textbf{Model}} &
\multicolumn{2}{c}{\textbf{Single-finger}} &
\multicolumn{2}{c}{\textbf{Multi-finger}} &
\multicolumn{2}{c}{\textbf{Wrist}} &
\multicolumn{2}{c}{\textbf{Rest}} &
\multicolumn{2}{c}{\textbf{Overall}} \\
\cmidrule(lr){2-3} \cmidrule(lr){4-5} \cmidrule(lr){6-7} \cmidrule(lr){8-9} \cmidrule(lr){10-11}
& \textbf{ACC} & \textbf{STD}
& \textbf{ACC} & \textbf{STD}
& \textbf{ACC} & \textbf{STD}
& \textbf{ACC} & \textbf{STD}
& \textbf{ACC} & \textbf{STD} \\
\midrule
w/o Iteration       & 78.25 & 0.02 & 79.34 & 0.03 & 80.05 & 0.02 & 80.72 & 0.01 & 79.59 & 0.02 \\
w/o Warm-up         & 87.20 & 0.02 & 88.94 & 0.02 & 92.45 & 0.02 & 93.37 & 0.01 & 91.65 & 0.01 \\
Full Model (Ours)   & \textbf{92.25} & 0.02 & \textbf{94.50} & 0.01 & \textbf{95.19} & 0.01 & \textbf{96.82} & 0.01 & \textbf{95.14} & 0.01 \\
\bottomrule
\end{tabular}%
}
 \vspace{-2mm}
\end{table}

\noindent\textbf{Effect of Human-Linguistic Biases.}
We further evaluate the effectiveness of the human-linguistic biases, including the Zipf’s law prior, the context-sensitivity constraint, and the inter-generational preserving bias. As shown in Table~\ref{tab:ablation_human}, removing all these biases (``w/o Human Bias'') leads to a substantial performance drop, with overall accuracy decreasing from 95.14\% to 87.35\%. When individual biases are removed, the accuracy also declines, but to different extents: eliminating the Zipf prior (``w/o Zipf Bias'') reduces the model’s ability, while removing the context-sensitivity constraint (``w/o Context Bias'') weakens the local semantic coherence of the emergent language. In addition, excluding the inter-generational preserving bias (``w/o Preserving Bias'') slightly affects performance, indicating that this term mainly stabilizes context inheritance rather than directly improving accuracy. The full model achieves the highest accuracy. Overall, these results confirm that human linguistic biases help the model form natural language-like representations, thereby facilitating the emergence of ``sEMG language.''

\begin{table}[t]
\centering
\vspace{-3mm}
\caption{Effect of human-linguistic biases in terms of ACC (\%) and STD.}
\vspace{-4mm}
\label{tab:ablation_human}
\setlength{\tabcolsep}{4pt}
\small
\resizebox{\columnwidth}{!}{%
\begin{tabular}{lcccccccccc}
\toprule
\multirow{2}{*}{\centering\textbf{Model}} &
\multicolumn{2}{c}{\textbf{Single-finger}} &
\multicolumn{2}{c}{\textbf{Multi-finger}} &
\multicolumn{2}{c}{\textbf{Wrist}} &
\multicolumn{2}{c}{\textbf{Rest}} &
\multicolumn{2}{c}{\textbf{Overall}} \\
\cmidrule(lr){2-3} \cmidrule(lr){4-5} \cmidrule(lr){6-7} \cmidrule(lr){8-9} \cmidrule(lr){10-11}
& \textbf{ACC} & \textbf{STD}
& \textbf{ACC} & \textbf{STD}
& \textbf{ACC} & \textbf{STD}
& \textbf{ACC} & \textbf{STD}
& \textbf{ACC} & \textbf{STD} \\
\midrule
w/o Human Bias       & 85.75 & 0.03 & 87.42 & 0.03 & 87.93 & 0.02 & 88.20 & 0.02 & 87.35 & 0.02 \\
w/o Zipf Bias        & 89.51 & 0.02 & 90.26 & 0.02 & 92.45 & 0.02 & 93.18 & 0.02 & 90.65 & 0.02 \\
w/o Context Bias     & 89.60 & 0.02 & 91.42 & 0.02 & 92.75 & 0.01 & 94.35 & 0.01 & 92.13 & 0.02 \\
w/o Preserving Bias  & 90.30 & 0.02 & 92.56 & 0.01 & 94.28 & 0.01 & 95.47 & 0.01 & 93.85 & 0.01 \\
Full Model (Ours)    & \textbf{92.25} & 0.02 & \textbf{94.50} & 0.01 & \textbf{95.19} & 0.01 & \textbf{96.82} & 0.01 & \textbf{95.14} & 0.01 \\
\bottomrule
\end{tabular}%
}
\vspace{-1mm}
\end{table}

\noindent\textbf{Effect of Residual-Based Adaptation.}
We ablate the residual-guided adaptive token allocation strategy, which allocates tokens based on the local residual energy of sEMG signals. As shown in Table~\ref{tab:ablation_residual}, removing this mechanism (``w/o Residual-based'') lowers accuracy from 95.14\% to 92.60\%. The accuracy reduction is larger for dynamic gestures, with reductions of 2.50\% for single-finger and 3.30\% for multi-finger activities, and smaller for steady activities, with reductions of 1.45\% for wrist and 0.95\% for rest. These results confirm that adaptive tokenization better preserves the dynamic characteristics of sEMG signals and enables the ``sEMG language'' to produce more discriminative representations.

\begin{table}[t]
\centering
\caption{Effect of residual-based adaptation in terms of ACC (\%) and STD.}
\vspace{-4mm}
\label{tab:ablation_residual}
\setlength{\tabcolsep}{4pt}
\small
\resizebox{\columnwidth}{!}{%
\begin{tabular}{lcccccccccc}
\toprule
\multirow{2}{*}{\centering\textbf{Model}} &
\multicolumn{2}{c}{\textbf{Single-finger}} &
\multicolumn{2}{c}{\textbf{Multi-finger}} &
\multicolumn{2}{c}{\textbf{Wrist}} &
\multicolumn{2}{c}{\textbf{Rest}} &
\multicolumn{2}{c}{\textbf{Overall}} \\
\cmidrule(lr){2-3} \cmidrule(lr){4-5} \cmidrule(lr){6-7} \cmidrule(lr){8-9} \cmidrule(lr){10-11}
& \textbf{ACC} & \textbf{STD}
& \textbf{ACC} & \textbf{STD}
& \textbf{ACC} & \textbf{STD}
& \textbf{ACC} & \textbf{STD}
& \textbf{ACC} & \textbf{STD} \\
\midrule
w/o Residual-based & 89.75 & 0.02 & 91.20 & 0.02 & 93.74 & 0.01 & 95.87 & 0.01 & 92.60 & 0.01 \\
Full Model (Ours)  & \textbf{92.25} & 0.02 & \textbf{94.50} & 0.01 & \textbf{95.19} & 0.01 & \textbf{96.82} & 0.01 & \textbf{95.14} & 0.01 \\
\bottomrule
\end{tabular}%
}
\vspace{-5mm}
\end{table}

\section{Conclusion}
\label{sec:conclusion}
In this work, we present LLM-sEMG, which enables large language models to perform accurate activity recognition from sEMG signals. We design an sEMG-oriented VQ-VAE to facilitate the emergence of an ``sEMG language,'' transforming continuous sEMG signals into human-like linguistic representations. By integrating human linguistic biases with a residual-based adaptive token allocation strategy, the evolution of the ``sEMG language'' is facilitated and its discriminative expressiveness is enhanced. Experiments on the GRABMyo and NinaPro~DB2 datasets demonstrate that LLM-sEMG achieves superior performance over existing baselines

{
    \small
    \bibliographystyle{ieeenat_fullname}
    \bibliography{main}
}


\end{document}